# Image retrieval approach based on local texture information derived from predefined patterns and spatial domain information

Nazgol Hor[1], Shervan Fekri-Ershad[1,2,*]

[1] Faculty of computer engineering, Najafabad Branch, Islamic azad university, Najafabad, Iran
[2] Big data research center, Najafabad Branch, Islamic azad university, Najafabad, Iran
*fekriershad@pco.iaun.ac.ir

**Abstract:** With the development of Information technology and communication, a large part of the databases is dedicated to images and videos. Thus retrieving images related to a query image from a large database has become an important area of research in computer vision. Until now, there are various methods of image retrieval that try to define image contents by texture, color or shape properties. In this paper, a method is presented for image retrieval based on a combination of local texture information derived from two different texture descriptors. First, the color channels of the input image are separated. The texture information is extracted using two descriptors such as evaluated local binary patterns and predefined pattern units. After extracting the features, the similarity matching is done based on distance criteria. The performance of the proposed method is evaluated in terms of precision and recall on the Simplicity database. The comparative results showed that the proposed approach offers higher precision rate than many known methods.

## I. INTRODUCTION

In big databases, huge amounts of data are stored in different formats such as text, images, videos, audio, etc., in a fraction of time. Therefore, big databases with the capacity to store this large amount of data are needed. So, image retrieval becomes an important field for research. Image retrieval means extracting the most similar images to the query (user-submitted image to the search engine) from a huge image database. In image retrieval problem, retrieval precision is the main to evaluate the efficiency of a system. Other challenges, such as noise, rotation, runtime, and scale variations, also affect on efficiency too. There are two strategies for image retrieval, text-based and content-based. Text-based image retrieval systems (TBIRs) use around text of image, tags, attached keywords to retrieve most similar images. These methods have been lost in use today for some reasons such as high costs, human complexity for text tagging, difficulty to define image contents in text format and lack of texts for some image types such as web pages or mobile applications. The second group, commonly called content-based image retrieval (CBIR) systems. These methods use visual content features within an image including objects, color, texture, background, etc. The method presented in this paper can be categorized as a CBIR method. So, TBIR methods are not mainly in the field of this paper, and will no longer be discussed. Nearly all of the CBIR systems have a same general flowchart which is shown in figure 1. In some cases a feedback edge is also used to improve retrieval accuracy for learner systems. Feature extraction plays the main role in CBIR systems. Discriminative features, that can be a good representation of image contents, should be extracted from all images within the database using image descriptors. Then query's features are extracted using same descriptors. The similarity amount between the query's features with each database images is measured. Finally the most similar images from database are presented to the user. In this process, image descriptors play key role. If the descriptors extract more discriminative features, the system's performance will be increased. There are various methods for CBIR so far. The main difference is being in the feature extraction or similar matching phase. Researchers usually try to employ descriptors that extract features related to the texture, color, or shape of the image. Our researches show that texture, shape, and color features cannot be good representations of the content and objects in the image individually. Therefore, this paper attempts to develop a method that extracts, locally and globally, texture information jointly. In this respect, local spatial structure and texture information are extracted using combination of evaluated local binary pattern (ELBP) and predefined pattern units (Pre-PU). In order to consider the color information, input colored image is separated to color channels. Finally, for the similarity matching step, some efficient similarity/distance metrics have been employed to select the most compatible distance measure with the extracted features. The performance of the proposed approach is evaluated in terms of precision and recall rate, on benchmark Simplicity dataset. The performance of the proposed approach has been compared with several well-known methods in this field, under the same validation conditions, which can be seen in the results section. The comparative results showed that the proposed approach provides better precision and recall rates than compared methods.

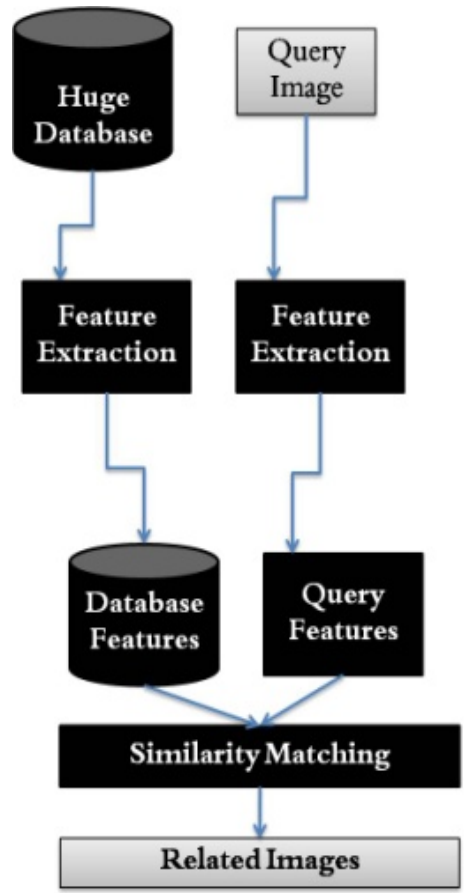

Fig. 1. General flowchart of most content based image retrieval systems

*A. Paper organization*

The paper is organized as follows: In section II, some related and well known CBIR systems are discussed to analysis motivations of this paper. In section III, our proposed approach is proposed which is combination of evaluated local binary patterns and predefined pattern units. Section IV presents the experimental results. Finally, the conclusion is included.

## II.  RELATED IMAGE RETRIEVAL APPROACHES

Image retrieval systems work by one of two strategies, text-based or content-based. Text-based methods, for the reasons mentioned above are not popular today. So, in this section, only some efficient CBIR systems are survived with details. As noted above, extracting texture and color information separately cannot well define the content of an image, so combination CBIR approaches have become increasingly popular in the last decade.

For example, In [1], Tajeripour et al. presented a CBIR approach based on early fusion of color and texture features. In [1], tajeripour et al., employ all components of color space. In [1], the input image is first transferred to RGB color space. Then color channels are separated. Evaluated local binary pattern (ELBP) operator is applied to each channel individually. Three numeric feature vectors with 3P+6 dimensions are extracted, where P is the number local neighborhoods. Also, the variance histogram is considered as another feature set. Finally, by concatenating the extracted features, a final feature vector is generated that is used for similar matching phase. The logarithmic likelihood metric is used in the similar matching phase. In [1], a pre-processing step, called object cropping is used based on morphological operations. The preprocess step can be useful for images with an index object and can result in deletion of the background. Consideration of texture and color information simultaneously and the good ability of the ELBP to extract discriminative features, are some advantages of [1]. However, the performance is greatly reduced for images that contain more than one index object. It just analyzes local information and does not consider global information.

In [2], singh and pritkar have proposed a method for image retrieval that is faster than many existing methods. The higher retrieval speed in [2] is because of using reduction algorithm in feature extraction phase. The authors in [2] claim that texture and color analysis only in one component of a color space yields similar results to the analysis of these features across all components of a color space. Hence, it has a much lower computational complexity. In this regard, in [2], the brightness component is first measured in a colored RGB image based on the formula $I = R + G + B / 3$ and then three operations such as color histogram, block differences of invariant probabilities (BDIP) and block variation of local correlation coefficients (BVLCC) is applied just in brightness channel to extract feature. In [2], it is discussed that component V in HSV color space, component L in color space La * b *, and channel Y in YCbCr color space can also play the role of brightness component.

In [3] Rangkuti et al., proposed a fuzzy based approach for batik motif image retrieval. In this respect, texture and shape features are employed in a fusion manner to achieve best result. Also, canbera is used in [3] as similarity metric. Zheng et al. [19] have proposed a block-based method for image retrieval. In feature extraction phase, the image is first transferred to the HSI color space. And then is divided into sufficient blocks with overlaps. Next, the main block is selected. Finally, two vertical and horizontal histograms are generated after edge tracking with roberts. In [19], the weighted Euclidean distance metric is used for similarity matching phase. The weights have been determined experimentally and no specific study has been done on their arrangement. Low accuracy compared to some of the existing methods and non-adjusting weights are the obvious drawbacks of [19]. An approach is proposed by pawar and belagali in [4] based on texture information extracted using basic local binary patterns (LBP). Some limitation of basic LBP such as noise sensitivity and color information disregarding can be seen in [4]. The method presented in [4], doesn't provide high accuracy on multi-object images in comparison with well known CBIR systems.

Deep convolution networks are widely used today for image retrieval. Gokso et al., developed a CBIR system in [5] using features extracted from deep convolution network. The input material of CNN in [5] is intensity and coordination information of the image pixels. Network weights are optimized to provide discriminative output features. A feature reduction algorithm is used in [5] to reduce runtime. High accuracy in comparison with some of the exciting methods and rotation insensitivity are some advantages of [5]. However, the high computational complexity and noise sensitivity, due to cross-sectional adjustment of weights, are weaknesses points of [5].

Sharma et al., [6] used color histogram for the feature extraction step. This method is fast enough for online applications. But, retrieval accuracy of [6] is not high because of using global color information only. Sensitivity to noise and scale variations are other disadvantages of using pure color information. In [7], image is analyzed in both spatial and frequency domains. In [7], the quantized color histogram and regional wavelet features are used to describe image contents. The main novelty of [7] is using C-means algorithm for image segmentation, which finally results local wavelet features. Using frequency information improves the accuracy of this system in comparison with some other methods. However, the run time of [7] is high. It is sensitive to C-Means input parameters and the number of main objects in the image. In [23], bani and fekri-ershad proposed a CBIR approach based on combination of global and local features in two spatial and frequency domains. In this way, first image is filtered by Gaussian filter, then co-occurrence matrices are made in different directions. Next a selected sub set of statistical features are extracted from GLCMs. As a set of color features, the quantized color histogram is employed to extract global color information in spatial domain. To evaluate better performance, Gabor filter banks are used to extract local texture information in the frequency domain. Noise-resistant and rotation-invariant can be mentioned as advantages of [23]. High complexity because of using different feature types is one of the main disadvantages of [23]. The main difference between CBIR approaches is in the feature extraction algorithm, but most of the researchers believe that texture and color information should be extracted jointly. Lin et al. [8] used adaptive motif co-occurrence matrix (AMCOM) and matched color histogram in feature extraction phase. In [9], color information is extracted based on Zernik-like moment descriptor and texture information using steerable pyramid decomposition (SPD) operation. In [10] texture and color information was analyzed in the form of three histograms of color, distance, and angle. Yue et al., in [11] proposed a multi step CBIR system. Images first transferred to the HSV color space and then blocked. Each color block histogram is extracted separately. The intensity range is quantified to 10 bins and the overall color histograms of the image are also extracted. Also co-occurrence matrix is used in [11] to achieve texture information. In the similarity matching phase, each set of vectors is weighted and the most similar images of the data base are ranked. Image optimization is the process of enhancing the image quality in order to provide preferable transfer representation for many applications such as content based image retrieval systems. There are many state-of-art algorithms in this scope such as multi-objective scheduling cuckoo optimization algorithm [12]. Spectrum histogram is used for feature extraction phase in [13]. Finally energy amount of the produced histograms is used as the criterion for similarity matching phase. The proposed CBIR system in [14], the image is divided into valuable spaces with the help of Gabor filtering. Combination of zernick moments, gabor filter and color histogram are used in feature extraction phase. Dynamic CBIR system is considered as a new hot research topic research computer vision applications. Dynamic CBIR systems usually use learning models which can be retrained with user feedback. A dynamic CBIR system is proposed by Tzelepi and Tefas [29], based on learning convolutional representations of database images. A deep convolutional neural network model was employed to obtain the feature representations from the activations of the convolutional layers using max-pooling. Then three basic model retraining approaches is suggested to consider user feedbacks. The fully unsupervised retraining, the retraining with relevance information and the relevance feedback based retraining. These three retraining models make it more useful in real applications, especially enough feedback is provided. But, the lack of enough training or feedback information decreases the performance of the method. For further study about this scope, a good review on CBIR systems has been done in [15].

### III. THE PROPOSED IMAGE RETRIEVAL APPROACH

The content of an image is recognizable by visual components such as texture, color and shape in human visual system. Therefore, to provide an accurate CBIR system, texture and color information must be extracted simultaneously. Some researches in this field also showed that examining texture and color information in local or global individually cannot be enough. So, in this paper, a CBIR approach is proposed which extracts texture information in global and local manners. There are usually two strategies to combine texture and color features, namely early fusion and late fusion [16]. In early fusion, texture descriptors are applied to each color channel separately, and finally a set of composite features is extracted. These features provide high discrimination power when texture and color information are blended into the pixel level of the image. In late fusion, the texture and color features are combined across the entire image surface. In this strategy, two set of features are extracted with two different operations (in the form of feature vectors or histograms), and finally these two sets are joined together to form the final representation. The proposed approach use early fusion strategy to combine color and texture information.

*A. Evaluated local binary patterns (ELBP)*

Local binary pattern (LBP) was first proposed by Ojala et al. [17] for image texture analysis. The results in [17] showed that LBP can extract local spatial structure information and local contrast of the image. To use this operator, first a neighborhood with radius *R* and the number of neighborhood *P* should be considered for each pixel. Then the intensity of each neighbor's is compared with the neighboring central. A binary code is provided for each local neighborhood with *P* bits. The value one is assigned to $k^{th}$ bit, if the $k^{th}$ neighbor's intensity be more than central pixel intensity value. The value zero is assigned to $k^{th}$ bit, if the $k^{th}$ neighbor's intensity be less than central pixel value. For example, if the neighborhood radius considered as one, 8 neighbors will be defined and the extracted binary pattern will be in eight bits. The output of this pattern based on the decimal format must be a number between 0 to 255. In [17], the histogram of the LBP values in whole image is considered as features. Eq. (1) shows how to calculate basic LBP. An example of LBP calculation process is also shown in Figure 2.

$$\mathbf{LBP}_{P,R}(c) = \sum_{k=1}^{p} \delta(I_k - I_c) 2^{k-1} \quad (1)$$

$$\boldsymbol{\delta}(x) = \begin{cases} 1 & \text{if} \quad x \geq 0 \\ 0 & \text{else} \end{cases} \quad (2)$$

Where, $I_c$ is the intensity value of central pixel of neighborhood and $I_k$ shows the intensity value of the $k^{th}$ neighborhood. *P* shows the total number of neighbors, which has a direct relation with the radius *R*.

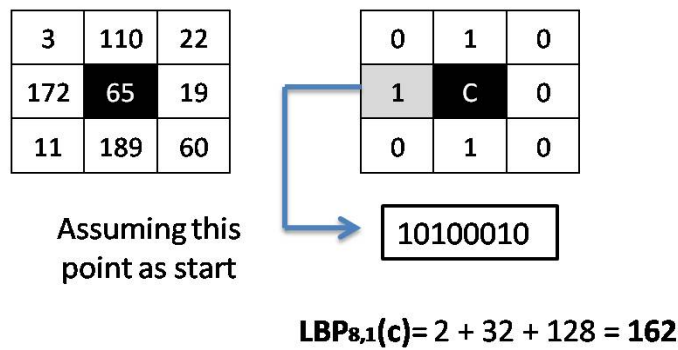

Fig. 2. An example of LBP calculation process with R=1 and P=8

The output histogram of the basic LBP is highly sensitive to the starting point of the binary pattern producing (matrix of weights). Also the results of [18] show that the above basic LBP doesn't provide high discriminative features and its computational complexity is high. Therefore, Ojala et al., [18] presented an evaluated version of LBP, abbreviated as ELBP. The ELBP process is similar to the LBP in binary pattern providing. But the feature extraction method is different. In ELBP, after generating the binary pattern (no matter the starting point), a criterion called the uniformity amount (U) is calculated. Uniformity measure is defined as number of bitwise transactions between zero and one in the binary pattern. Then, based on the uniformity threshold, usually considered P/4 [19, 20], the binary patterns are divided into two groups: uniform and non-uniform. Uniform patterns are labeled based on the number ones in extracted binary pattern, and all non-uniform patterns are also labeled as P+1. How to calculate the uniformity and labeling of ELBP is shown in the following equation.

$$U(LBP_{P,R}(c)) = |\delta(I_1 - I_c) - \delta(I_p - I_c)| + \sum_{k=2}^{P}|\delta(I_k - I_c) - \delta(I_{k-1} - I_c)| \quad (3)$$

$$ELBP_{P,R}^{riu_T}(c) = \begin{cases} \sum_{k=1}^{P}\delta(I_k - I_c) & \text{if } U(LBP_{P,R}(c)) \leq U_T \\ P + 1 & \text{elsewhere} \end{cases} \quad (4)$$

By applying the ELBP operator, each pixel will have a label between zero to P+1. Finally, the occurrence probability of each label is considered as a feature. So, a feature vector can be extracted with P+2 dimensions.

*B. Predefined pattern units (Pre-PU)*

The predefined pattern unit (Pre-PU) operator is one of the image texture analysis operations that can extract repetitive texture information based on edge structures. It was first introduced in [21] to classify stone textures. It has been used by researchers in other applications of image analysis such as texture classification [22], visual inspection systems [30], etc. To apply the Pre-PU, first edge detection is done and a binary image is provided. Then predefined patterns in *n×m* size are detected and labeled in the image. Finally, the occurrence probability of predefined labels is considered as feature. The Pre-PU extracting process steps are as follows:

1. Edge detection operation is applied on the input image
2. A square/circular neighborhood is considered around each pixel
3. If the central pixel value is one, the total number ones in the neighbors is considered as label value
4. If the central pixel value is zero, regardless of the neighbors, the Pre-PU value of the neighborhood is considered as zero.

Some examples of local neighborhoods with different Pre-PU values are shown in the Figure 3. As can be seen, the Pre-PU operator, unlike the ELBP, follows the occurrence of some predefined patterns in edge detected image. But, the ELBP defines binary patterns based on local variance differences. Pre-PU are able to analyze continuous linear and non-linear repeating textures in the image. However, ELBP just analyzes local spatial structures and local variance.

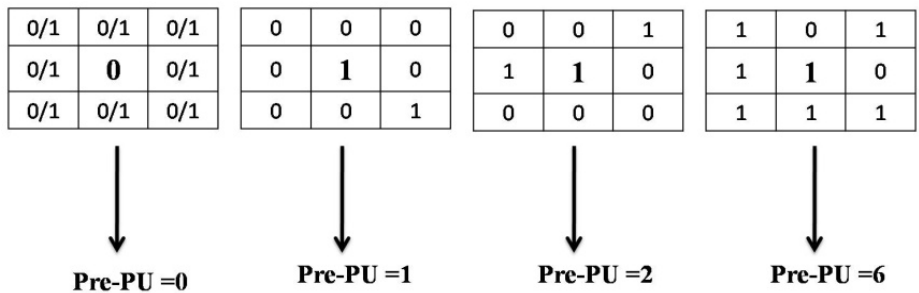

Fig. 3. Some examples of neighborhoods with different Pre-PU value after edge detection

*C. Proposed texture information fusion algorithm*

The main purpose of this paper is to propose a CBIR approach based on combination of color and texture information. So far, four different operations ELBP, and Pre-PU and color quantized histograms have been proposed for the feature extraction phase. The early fusion strategy is used in this paper to combine texture and color features. In this respect, first the image is transferred to the RGB color space. Then color channels are separated. Finally, ELBP and Pre-PU operators are applied to each color channel.

The number of extracted features using ELBP and Pre-PU are depends to neighborhood radius. For example, if the radius is one, the number of neighbors in each neighborhood will be eight. Therefore, a feature vector with 28 dimensions is being provided (ten ELBP features and nine Pre-PU features). The color histogram also depends to the quantization ratio in each color channel. All of the operations are applied to each color channel separately. So if we consider the number of neighbors P, the total number of dimensions in the final feature vector can be calculated by the following equation.

$$3 \times [(P+2) + (P+1)] = 6P + 9 \quad (5)$$

In online CBIR systems, the feature extraction phase for database is performed offline, and the extracted vectors are maintained in the database. As soon as the query image is presented by user, the feature extraction phase is performed just for query image. Therefore, the performance of our proposed CBIR approach does not decrease in online applications.

*D. Similarity matching phase*

As described above, CBIR systems usually have two main phases, feature extraction and similarity matching. In the similarity matching phase, the query content is compared with all database images individually. The most similar (minimum distance) database images to the query are sorted and displayed based on a similarity/distance metric. At this phase, a variety of distance metrics can usually be used, but selecting the most compatible metric with extracted features may increase the retrieval accuracy. In this paper, several distance metrics such as cosine, Euclidean, logarithmic likelihood ratio, city block, etc were tested and finally the Euclidean metric was provided the highest retrieval accuracy. How to calculate the Euclidean metric for the extracted feature vector in this paper is shown in the following equation.

$$E(Q, D_i) = E(Q_{red}, D_{i_{red}}) + E(Q_{green}, D_{i_{green}}) + E(Q_{blue}, D_{i_{blue}}) \quad (6)$$

Where

$$E(Q_k, D_{i_k}) = \sqrt{\sum_{j=0}^{P+1}(Q_{ELBP_j} - D_{i_{ELBP_j}})^2} + \sqrt{\sum_{j=0}^{P}(Q_{Pre-PU_j} - D_{i_{Pre-PU_j}})^2} \quad (7)$$

Where, Q is the query image and $D_i$ shows the i[th] image in database. P is the number of neighbors and k shows the color channel.

## IV. RESULTS

Two measures, precision and recall, are commonly used to measure the effectiveness and performance of CBIR systems. Precision is the ratio of the number of related retrieved images to the total number of retrieved images (the number of images submitted to the user per query image). Recall, is the ratio of the number of related retrieved images to the total number of related images in database. How to calculate these two criteria is shown in the following equations.

$$Precision = \frac{RR}{N} \times 100 \quad (8) \quad Recall = \frac{RR}{M} \times 100 \quad (9)$$

Where, RR is the number of related retrieved images, N is the total number of retrieved images and M is total number of related images in database. The related samples are images with a same class.

Simplicity [24], is a benchmark dataset which is used to evaluate the performance of the proposed CBIR approach. The Simplicity database contains 1000 images from 10 different classes titled Africa, beach, buildings, buses, dinosaurs, elephants, flowers, horses, mountains and food. Each class contains 100 different images. Some examples of this database are shown in Figure 4. Our experiments were performed using a Core i3 processor, 4 GB of RAM and 64-bit operating system with the help of Matlab 2016 and Python 3.7.

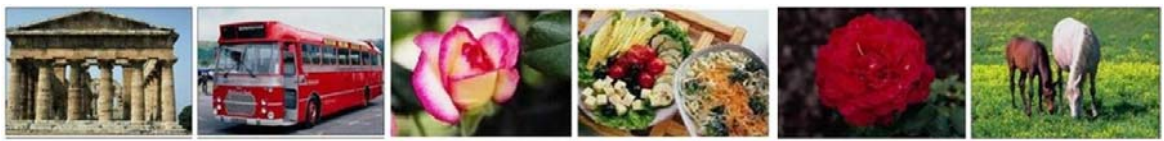

Fig. 4. Some examples of simplicity database

*A. Tune the installation parameters*

As discussed above, the proposed CBIR approach uses a combination of texture and color information extracted from four different local and global descriptors. Therefore, some input parameters of the operators have to be adjusted to achieve best performance. The proposed approach is an automatic system which should be used in different conditions. To provide fair comparisons, first we tuned the parameters on simplicity dataset with N=10 (number of retrieved images per query). Next, performance of the proposed approach with optimum parameters is compared with state-of-the-art methods on simplicity.

*A.a. Neighbor's radius*

In ELBP and Pre-PU descriptors, the neighborhood radius (R) have to be determined because it has a direct impact on the results and the number of extracted features. At this stage, the precision rate of the proposed method under the three different radius R=1, R=2 and R=3 for the above descriptors was evaluated on simplicity dataset. The results are shown in table 1. As can be seen in table 1, the radius R=1 provides the highest precision.

Table 1. Precision rate (%) based on different radius of ELBP and Pre-PU on simplicity dataset and N=10

| Neighbourhood radius | Precision rate (%) |
|---|---|
| R=1 | **82.95** |
| R=2 | 81.7 |
| R=3 | 79.5 |

*A.b. Distance metric*

Selecting the most consistent distance metric with extracted features can improve the proposed CBIR performance. For this purpose, the precision rate of our proposed CBIR approach is evaluated based on four different distance metrics such as cosine, Euclidean, logarithmic likelihood and city block. The results are shown in Table 2. As can be seen, the Euclidean metric provided the most precision rate. In all experiments, the neighborhood radius of the operators was set to R=1.

Table 2. Precision rate (%) based on different distance metrics in similarity matching phase on simplicity dataset and N=10

| Distance metric | Precision rate (%) |
|---|---|
| Canbera | 79.6 |
| Cosine | 81.78 |
| Euclidean | **82.95** |
| Log likelihood | 82.3 |
| City block | 79.04 |

*B. Comparison with state-of-the-art methods*

There are various CBIR approaches so far. At this point, we have compared the performance of our proposed CBIR approach with some state-of-the-art methods in this area. In the first experiment, the proposed approach is compared with some efficient methods in terms of precision and recall, evaluated on simplicity database with retrieve rate of N = 10, N = 20, N = 40. Results are shown in Tables 3. Due to the same validation conditions, the results reported in [2] and [25] are presented in table 4. Also, the presented CBIR systems in [1] and [4] were implemented and the results are listed below.

Table 3. Comparison with state-of-the-art methods in terms of recall and precision rate (%) on Simplicity

| Recall rate (%) | | | |
|---|---|---|---|
| Approach | Number of retrieved images | | |
| | N=10 | N=20 | N=40 |
| MLBP + Variance[1] | 8.1 | 15.2 | 26.8 |
| Texture information of brightness color component [2] | 8.32 | 15.2 | 28.9 |
| Basic LBP [4] | 8.0 | 14.8 | 26.0 |
| Texture + Shape characteristics [3] | 8.12 | 15.36 | 26.77 |
| Multi texton histogram [26] | 8.0 | 14.5 | 26.3 |
| Gabor + Co-occurrence matrix + Color histogram [23] | **8.60** | **16.58** | 30.51 |
| Proposed approach | **8.29** | **16.02** | **30.52** |
| Precision rate (%) | | | |
| Approach | Number of retrieved images | | |
| | N=10 | N=20 | N=40 |
| MLBP + Variance[1] | 81.33 | 76.22 | 67.22 |
| Texture information of brightness color component [2] | 83.2 | 76.37 | 72.29 |
| Basic LBP [4] | 80.05 | 74.37 | 65.04 |
| Texture + Shape characteristics [3] | 81.27 | 76.84 | 66.94 |
| Multi texton histogram [26] | 80.68 | 72.86 | 65.86 |
| Gabor + Co-occurrence matrix + Color histogram [23] | **86.04** | **82.94** | 76.27 |
| Proposed approach | **82.95** | **80.12** | **76.30** |

As can be seen in the table 3, our proposed approach provides higher precision and recall rate than all of the compared methods. The method presented in [23] uses a combination of texture and color information in global and local manner. Therefore, the precision of [23], in the case of retrieving 10 and 20 images is higher than our presented approach. However, as the number of retrieval images is increased, the precision and recall rate of [23] falls sharply in comparison with our proposed approach. Due to the same validation conditions, the results reported in [2] and [26] are presented for comparison. The CBIR systems presented in [1], [4] and [23] are also implemented. As can be seen in the tables 3, our proposed approach provides higher precision and recall rate than near all of the compared methods.

## V. CONCLUSION

Image retrieval is a most important research topic in computer vision today. In this respect, an approach is proposed in this paper as combination of two evaluated texture descriptors. The texture information of the image extract locally and globally in the proposed approach. These features combined with color information based on early fusion strategy, to increase the performance. Thus the ELBP and Pre-PU were applied to each one of the color channels individually. Finally, the extracted feature vectors were normalized and concatenated. Comparative results showed that the proposed approach provides higher precision and recall rate than many efficient methods in this field. The computational complexity of the proposed system was calculated based on number of extracted features. The results showed that the computational complexity of our proposed approach is lower than many related methods in this scope. The effectiveness of the proposed method is due to combination of local and global information of texture and early fusion of color information. Rotation invariant, due to the resistance of Pre-PU and ELBP operators to these factors, are another advantages of the proposed method. Fusion of Pre-PU and ELBP as proposed in this paper is a general way to extract texture information of the input image. So, it can be used in many computer vision problems.